\documentclass{article}
\usepackage{ijcai17}

\usepackage{times}
\usepackage{epsfig}
\usepackage{graphicx}
\usepackage{amsmath}
\usepackage{amssymb}
\usepackage{mathsymb}
\usepackage{textcomp}
\usepackage{verbatim}
\usepackage{subfigure}
\usepackage{url}
\usepackage{graphicx}
\usepackage{multirow}
\usepackage[super]{nth}
\usepackage{balance}
\usepackage{colortbl}
\graphicspath{{./}{./Fig/}{./Figs/}{./Fig/eps/}{./Fig/pdf/}}
\usepackage[ruled,linesnumbered,vlined]{algorithm2e}

\newcommand{\etal}{\textit{et al.\ }}
\newcommand{\eg}{\emph{e.g.,\ }}
\newcommand{\ie}{\emph{i.e.,\ }}
\newcommand{\etc}{\emph{etc..\ }}

\title{Multi-View Spectral Clustering via Structured Low-Rank Matrix Factorization}
\author{Yang Wang$^{\dag}$ and Lin Wu$^{\ddag}$ \\
 $^{\dag}$The University of New South Wales, Kensington, Sydney, Australia\\
 $^{\ddag}$ The University of Queensland, Brisbane, Australia\\
 wangy@cse.unsw.edu.au, lin.wu@uq.edu.au
}

\begin{document}

\maketitle

\begin{abstract}
 Multi-view data clustering attracts more attention than their single view counterparts due to the fact that leveraging multiple independent and complementary information from multi-view feature spaces outperforms the single one.
Multi-view Spectral Clustering aims at yielding the data partition agreement over their local manifold structures by seeking eigenvalue-eigenvector decompositions. Among all the methods, Low-Rank Representation (LRR) is effective, by exploring the multi-view consensus structures beyond the low-rankness to boost the clustering performance.
However, as we observed, such classical paradigm still suffers from the following stand-out limitations for multi-view spectral clustering of (1) overlooking the flexible local manifold structure, caused by (2) aggressively enforcing the low-rank data correlation agreement among all views, such strategy therefore cannot achieve the satisfied between-views agreement; worse still, (3) LRR is not intuitively flexible to capture the latent data clustering structures.
In this paper,  we present the structured LRR by factorizing into the latent low-dimensional data-cluster representations, which characterize the data clustering structure for each view.  Upon such representation, (b) the laplacian regularizer is imposed to be capable of preserving the flexible local manifold structure for each view. (c) We present an iterative multi-view agreement strategy by minimizing the divergence objective among all factorized latent data-cluster representations during each iteration of optimization process, where such latent representation from each view serves to regulate those from other views, such intuitive process iteratively coordinates all views to be agreeable. (d) We remark that such data-cluster representation can flexibly encode the data clustering structure from any view with adaptive input cluster number.
To this end, (e) a novel non-convex objective function is proposed via the efficient alternating minimization strategy. The complexity analysis are also presented.  The extensive experiments conducted against the real-world multi-view datasets demonstrate the superiority over state-of-the-arts.
\end{abstract}

\section{Introduction}
Spectral clustering \cite{nips01,nips04,SKMEA,Fei11TNN} aims at exploring the local nonlinear manifold (spectral graph)\footnote{In the rest of this paper, we will alternatively use nonlinear manifold structure or spectral graph structure} structure \cite{FeiTNNLS15,FeiTNNLS16}, attracting great attention within recent years. With the development of information technology, multi-view spectral clustering, due to the fact of outperforming the single view counterparts by leveraging the complementary information from multi-view spaces. As implied by multi-view research \cite{TaoPAMIA,Taosurvey,ijcai16}, an individual view is not capable of being faithful for effective multi-view learning. Therefore, exploring multi-view information is necessary, and has been demonstrated by a wide spectrum of applications \eg similarity search \cite{Shao-TIP15,YLXSIGIR15,LYMM13,land4,LYJMMM13,YLXMM14,YLXMM15,YXLTIP17,YXQCIKM13}, human action recognition \cite{shao-cvpr14,Shao-IJCV16,LYGX17,YXLKAIS16,YXLPAKDD14,LYPR17,YLIVC17,YLCJ12} \etc

Essentially, given the complementary information from multi-views, the critical issue of multi-view clustering is to achieve the multi-view clustering agreement/consensus \cite{Taosurvey,TaoTIP14,iccv15nie} to yield a substantial superior clustering performance over the single view paradigm. Numerous multi-view based methods are proposed for spectral clustering. \cite{CVPR12,M-V-C,YLINS13} performs multi-view information incorporation into the clustering process by optimizing certain objective loss function. \emph{Early fusion} strategy can also be developed by concatenating the multi-view features into a uniform one \cite{cvpr10}, upon which the similarity matrix is calculated for further multi-view spectral clustering. As mentioned by \cite{YangMVC-TIP15,land3}, such strategy will be more likely to destroy the inherent property of original feature representations within each view, hence resulting into a worse performance; worse still, sometimes, as indicated by the experimental reports from our previous research \cite{YangMVC-TIP15}, it may even be inferior to the clustering performance with a single view. In contrast, \emph{Late fusion} strategy \cite{ecmlpkdd09} conducts spectral clustering performance for each view, and then combining multiple them afterwards, which, however, cannot achieve the multi-view agreement, without collaborating with each other.

Canonical Correlation Analysis (CCA) based methods \cite{CSC,ICML09} for multi-view spectral clustering project the data from multi-view feature spaces onto one common lower dimensional subspace, where the spectral clustering is subsequently conducted. One limitation of such method lies in the fact that such common lower-dimensional subspace cannot flexibly characterize the local manifold structures from heterogeneous views, resulting into an inferior performance. Kumar \etal \cite{NIPS11} proposed a state-of-the-art co-regularized spectral clustering for multi-view data. Similarly, a co-training \cite{COLT98,ICML10} model is proposed for this problem \cite{icml11}.

One assumption for above work \cite{NIPS11,icml11}  is the scenario with noise corruption free for each view. However, it is not easily met. To this end, Low-Rank Representation (LRR) \cite{RMVSC,LRRICML2010,YangMVC-TIP15,ICCV2011,LiuPAMI13} is proposed. As summarized in \cite{ijcai16}, where the basic idea is to decompose data representation into a view-dependent noise corruption term and a common low rank based representation shared by all views, leading to common data affinity matrix for clustering; The effectiveness of low-rank model also leads to numerous research on multi-view subspace learning \cite{AAAI16,ICDM144} applied to the pattern recognition field.


LRR tries a common multi-view low-rank representation, but overlooks the distinct manifold structures. To remedy the limitations, inspired by the latest development of  graph regularized LRR \cite{Jun-TPAMI16,Jun-TIP15}, we recently proposed another iterative views agreement strategy \cite{ijcai16} with graph regularized Low-rank Representation for multi-view spectral clustering, named \textbf{LRRGL} for short, to characterize the non-linear manifold structure from each view, \textbf{LRRGL} couples LRR with multi-graph regularization, where each one can characterize the view-dependent non-linear local data manifold structure \cite{Lin-Neuro16}. A novel iterative view agreement process is proposed of optimizing the proposed, where, during each iteration, the low-rank representation yielded from each view serves as the constraint to regulate the representation learning from other views, to achieve the consensus, implemented by applying Linearized Alternating Direction Method with Adaptive Penalty \cite{LinNIPS2011}.

Despite the effectiveness of \textbf{LRRGL}, we still identify the following non-trivial observations that are not addressed by \textbf{LRRGL} to obtain the further improvement
\begin{itemize}
\item It is less flexible for $Z_i$ yielded by low-rank constraint to capture the flexible latent data similarity that can encode the more rich similarity information than $Z_i$ over $X_i$; that can be better solved by matrix factorization.

\item \textbf{LRRGL} mainly focused on yielding the low-rank primal data similarity matrix $Z_i$ derived from $X_i$. However, such primal $Z_i$ is less intuitive to understand and less effective to reveal the ideal data clustering structure for the $i^{th}$ view, as well as multi-views. Hence, it will prevent that achieving the better multi-view spectral clustering performance. The structured consensus loss term imposed over $Z_i (i \in V)$ may not effectively achieve the consensus regarding the multi-view spectral clustering.
\end{itemize}

\begin{figure}[t]
\begin{tabular}{cc}
\includegraphics[width=4cm,height=4cm]{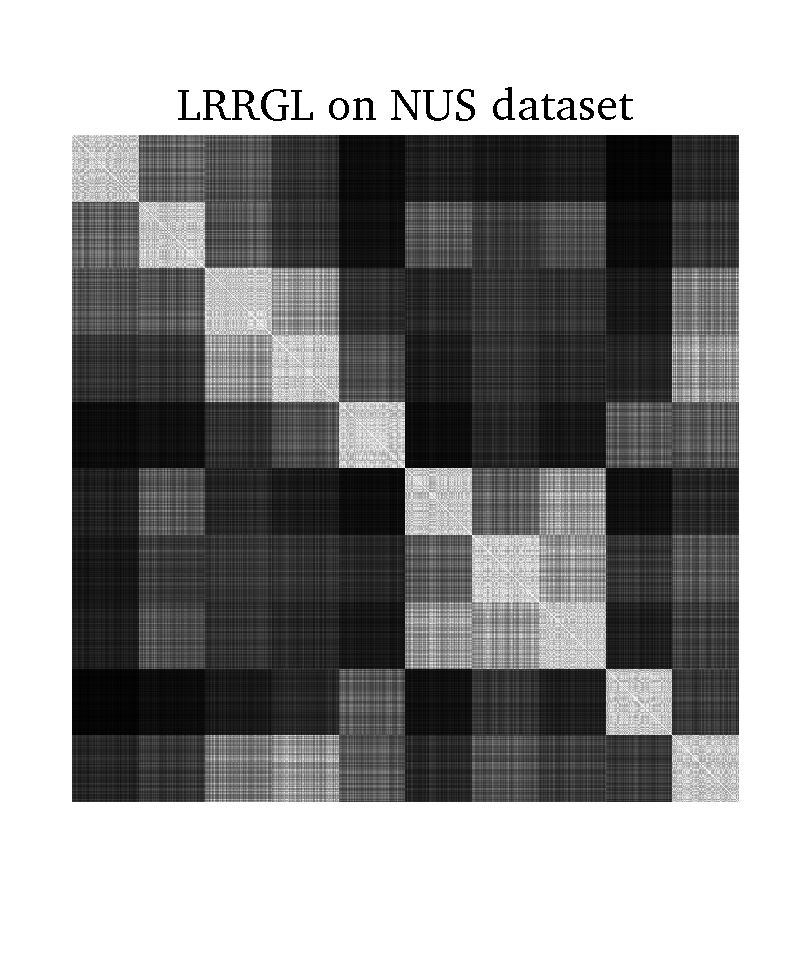}&
\includegraphics[width=4cm,height=4cm]{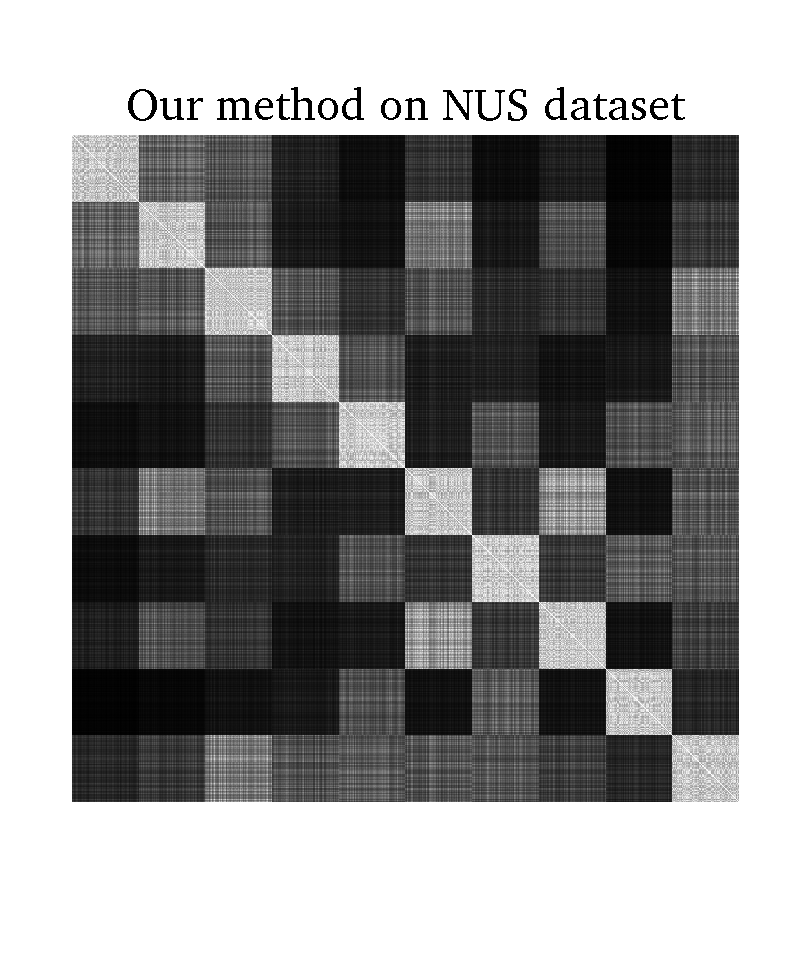}\\
(a)&(b)
\end{tabular}
\caption{The visualization results of the multi-view (please refer to section ~\ref{sec:exp} for specific multi-view features) affinity matrix between ours and \textbf{LRRGL} over NUS data; The more whiter for diagonal blocks, the more ideally the cluster is to characterize the data objects within the larger similarity, meanwhile, the more blacker for non-diagonal blocks, the more reasonable the non-similarity data objects are unlikely to cluster together. For such result,
we can see the diagonal blocks from $3^{rd}$ to the $8^{th}$ of our method are more whiter than those of \textbf{LRRGL}, leading to the result that the surrounding black non-diagonal blocks of our method are more salient than those of \textbf{LRRGL}, which demonstrate the advantages of our method via a latent factorized data-cluster representation over \textbf{LRRGL}}\label{fig:motivation}
\end{figure}

\subsection{Our Contributions}
This paper is the extension of our recent work \cite{ijcai16}, upon that,  we deliver the following novel contributions to achieve the further improvement over multi-view spectral clustering
\begin{itemize}
\item Instead of focusing on primal low-rank data similarity matrix $Z_i$ such that $i = 1,\ldots,V$, we perform a symmetric matrix factorization over $Z_i$ into the data-cluster indicator matrix, so that such latent factorization provides the better chance to preserve the ideal cluster structure besides flexible manifold structure for each view.

\item We impose the laplacian regularizer over factorized data-clustered representation to further characterize the nonlinear local manifold structure for each view. We remark that the factorized data-cluster matrix can effectively encode the clustering structure, we provide an example to illustrate this in Fig.\ref{fig:motivation}.  \emph{To reach the multi-view clustering agreement, we set the same clustering number for all views to the data-clustering representation for all views.}

\item We impose the consensus loss term to minimize the divergence among all the latent data-cluster matrix instead of $Z_i$ to achieve the multi-view spectral clustering agreement.

\item To implement all the above insights, we propose a novel objective function, and an efficient alternating optimization strategy together with the complexity analysis to solve the objective function; moreover, we deliver the intuitions of iterative multi-view agreement over the factorized latent data-cluster representation during each iteration of our optimization strategy, that will eventually lead to the multi-view clustering agreement.

\item Extensive experiments over real-world multi-view data sets demonstrate the advantages of our technique over the state-of-the-arts including our recently proposed \textbf{LRRGL} \cite{ijcai16}.
\end{itemize}
Recently another elegant graph based PCA method \cite{Fei17neurocomputing} is proposed spectral clustering with out-of-sample case. Unlike this effective technique, we study the multi-view case to address the effective consensus for spectral clustering.  We summarize the main notations in Table ~\ref{table:notations}.

\begin{table}[t]
\caption{The Notations Summarization}
\tiny
\begin{center}
\begin{tabular}{cc}
\hline\hline
Notations & Explanation \\
\hline
$X_i \in \mathbb{R}^{d_i \times n}$ & Feature Representation Matrix for the $i^{th}$ view.\\
$E_i \in \mathbb{R}^{d_i \times n}$ & Feature noise matrix for the $i^{th}$ view.\\
$Z_i \in \mathbb{R}^{n \times n}$ & Self-expressive similarity matrix for the $i^{th}$ view.\\
$U_i \in \mathbb{R}^{n \times d}$ & Data-cluster matrix for the $i^{th}$ view.\\
$W_i \in \mathbb{R}^{n \times n}$ & Data similarity matrix over $X_i$ for the $i^{th}$ view.\\
rank($A$) & The rank of the matrix $A$.\\
$A[k]$ & Updated matrix $A$ at the $k^{th}$ iteration.\\
$AB$ & Matrix Multiplication between  $A$ and $B$.\\
$n$ & The number of data objects.\\
$d_i$ & The dimension of the feature space for the $i^{th}$ view.\\
$||\cdot||^T$ & The matrix transpose.\\
$||\cdot||_F$ & Frobenius norm.\\
$||\cdot||_*$ & Nuclear norm.\\
$||\cdot||_1$ & $\ell_1$ norm of matrix seen as a long vector.\\
$(\cdot)^{-1}$ & The matrix inverse computation.\\
$I_d$ & Identity matrix with the size of $d \times d$.\\
$||\cdot||_2$ & $\ell_2$ norm of a vector.\\
Tr($\cdot$) & Trace operator over the square matrix.\\
$\langle\cdot,\cdot\rangle$ & inner product.\\
$V$ & The set of all views.\\
$C_k$ & The $k^{th}$ data cluster.\\
$|C_k|$ & The cardinality of $C_k$.\\
$|V|$ & Cardinality of the set V.\\
$A(l,\cdot)$ & The $l^{th}$ row of the matrix $A$.\\
$A(\cdot,m)$ & The $m^{th}$ column of the matrix $A$.\\
$A(l,m)$ & The $(l,m)^{th}$ entry of the matrix $A$.\\
\hline
\end{tabular}
\end{center}
\label{table:notations}
\end{table}

\section{Structured Low-Rank Matrix Factorization to Spectral Clustering}
We get started from each single view \eg the $i^{th}$ view as
\begin{equation}\label{eq:tech1}
\min_{Z_i, E_i} \frac{\theta}{2}||X_i - X_iZ_i - E_i||_F^2 + ||Z_i||_* + \beta||E_i||_1,
\end{equation}
where $\theta$ and $\beta$ are the trade-off parameters, as aforementioned, we always adopt $D_i$ to be $X_i$, so that $X_i$ can be decomposed as clean component $X_iZ_i$ and another corrupted component $E_i$ for the $i^{th}$ view. One can easily verify that rank$(X_iZ_i)$ $\leq$ rank$(Z_i)$, hence minimizing rank$(Z_i)$ is equivalent to bounding the low-rank structure of clean component $X_iZ_i$.

Now we are ready to deeply investigate $||Z_i||_*$ for the $i^{th}$ view. Following \cite{nuclearfactorize}, we reformulate the nuclear norm $||Z_i||_*$ as
\begin{equation}\label{eq:tech2}
||Z_i||_* = \min_{U_i,V_i,Z_i=U_iV_i^T}\frac{1}{2}(||U_i||_F^2+||V_i||_F^2),
\end{equation}
where $U_i \in \mathbb{R}^{n \times d}$ and $V_i \in \mathbb{R}^{n \times d}$; $d$ is always less than $d_i$ since high-dimensional data objects always characterize the low-rank structure.

\subsection{Notes regarding $U_i$ and $V_i$ for multi-view spectral clustering}
Before further discussing the low-rank matrix factorization, one may consider the following notes that the factorized $U_i$ and $V_i$ may need to satisfy in the context of both the \textbf{within-view} data structure preserving and \textbf{multi-view} spectral clustering agreement:
\begin{enumerate}
\item The low-rank data structure should be characterized by the factorized $U_i$ or $V_i$ for the $i^{th}$ view, especially to characterize the underlying data clustering structure.

\item The factorized latent factors should well encode the manifold structure for the $i^{th}$ view, which, as previously mentioned, is critical to the spectral clustering performance.

\item Either the row based matrix $U_i$  or column based matrix $V_i$ is considered to meet the above two notes? if so, which one? One may claim both to be considered, which, however, may inevitably raise more parameters to be tuned.

\item Not only the factorized latent low-dimensional factors \eg $U_i$ or $V_i$, should meet the above notes within each view \eg the $i^{th}$ view, but also need the same scale to unify all views to reach possible agreement.
\end{enumerate}
To address all the above notes, in what follows, we will present our technique of data-cluster based structured low-rank matrix factorization.
\subsection{Data-cluster (landmark) based Structured Low-Rank Matrix factorization}
We aim at factorizing $Z_i$ as an approximate symmetric low-rank data-cluster matrix
to minimize the reconstruction error
\begin{equation}\label{eq:ui}
\min_{Z_i}||X_i - X_iZ_i||_F^2,
\end{equation}
where we assume the rank of $Z_i$ is $k_i$, such that $k_i$ is related to the data cluster number for the $i^{th}$ view. As indicated by \cite{LiuPAMI13},  minimizing the Eq.\eqref{eq:ui} is equivalent to finding the optimal rank $k_i$ approximation relying on skinny singular value decomposition of $X_i = V\Sigma U^T$ to yield the following optimal solution
\begin{equation}\label{eq:ui2}
Z_i^* = U_iU_i^T,
\end{equation}
where $U_i \in \mathbb{R}^{n \times k_i}$, such that $k_i$ denotes the top $k_i$ principle basis of $X_i$. Here we follow the assumption in \cite{Chris_ding12} to see $k_i$ as the cluster number of data objects within the $i^{th}$ view, and the data-cluster symmetric matrix factorization has been widely adopted by the numerous existing research including semi-supervised learning \cite{land1,land2}, metric fusion \cite{land3} and clustering \cite{SKMEA}. We aim at solving the following equivalent low-rank minimization over $Z_i$ via the clustered symmetric matrix factorization below
\begin{equation}\label{eq:reform0}
||Z_i||_* = \min_{U_i,Z_i = U_iU_i^T}||U_i||_F^2,
\end{equation}
where we often minimize the following for derivative convenience with respect to $U_i$
\begin{equation}\label{eq:reform}
||Z_i||_* = \min_{U_i,Z_i = U_iU_i^T}\frac{1}{2}||U_i||_F^2
\end{equation}
\textbf{Remark.} Following Eqs. \eqref{eq:ui} and \eqref{eq:ui2}, we initialize the $U_i \in \mathbb{R}^{n \times k_i}$ via a k means clustering over $X_i$ and normalize $U_i(j,k) = \frac{1}{|C_k|}$ provided $X_i(\cdot,j)$ \ie the $j^{th}$ data object is assigned to $C_k$. By such normalization, all the columns of $U_i$ are orthonormal; moreover,  they are within the same magnitude so as to perform the agreement minimization. Furthermore, such factorization can well address the aforementioned challenges, it is worthwhile to summarize them below

\begin{itemize}
\item The data cluster structure can be well encoded by such low-rank data-cluster representation within each view. The setting $U_i = V_i$ can avoid the more parameters and importance weight discussion provided $U_i \neq V_i$.

\item More importantly, inspired by the reasonable assumption hold by all the multi-view clustering research \cite{NIPS11,icml11,M-V-C,YangMVC-TIP15}. As indicated by \cite{YangMVC-TIP15}, the ideal multi-view clustering performance is that the common underlying data clustering structure is shared by all the views; we naturally set all the $U_i$ with the same size by adopting the same value for $k_i = d(i = 1,\ldots,V)$ \ie the clustering number, upon the same data objects number $n$ for all views, so that the feasible loss functions can be developed to seek the multi-view clustering agreement with the same clustering number for all views.
\end{itemize}

For spectral clustering from each view, we preserve the nonlinear local manifold structure of $X_i$ via such low-rank data-cluster representation $U_i$ for the $i^{th}$ view, which can be formulated as
\begin{equation}\label{eq:spectral}
\begin{aligned}
& \frac{1}{2}\sum_{j,k}^{n}||u_j^i - u_k^i||_2^2W_i(j,k) \\
& = \sum_{j = 1}^{N}(u_j^i)^Tu_j^iH_i(j,j) - \sum_{j,k}^{N}(u_k^i)^Tu_j^iW_i(j,k) \\
& = \textmd{Tr}(U_i^TH_iU_i) - \textmd{Tr}(U_i^TW_iU_i) = \textmd{Tr}(U_i^TL_iU_i),
\end{aligned}
\end{equation}
where $u_k^i \in \mathbb{R}^{d}$ is the $k^{th}$ row vector of $U_i \in \mathbb{R}^{n \times d}$ representing the linear correlation between $x_k$ and $x_j(j \neq k)$ in the $i^{th}$ view;  $W_i(j,k)$ encodes the similarity between $x_j$ and $x_k$ for the $i^{th}$ view;  $H_i$ is a diagonal matrix with its $k^{th}$ diagonal entry to be the summation of the $k^{th}$ row of $W_i$, and $L_i = H_i - W_i$ is the graph laplacian matrix for the $i^{th}$ view.

Following \cite{ijcai16}, we choose Gaussian kernel to define $W_{jk}^i$
\begin{equation}\label{eq:gaussian}
W_i(j,k) = e^{-\frac{||x_j^i - x_k^i ||^2_2}{2\sigma^2}}.
\end{equation}

We aim to minimize the difference of low-rank based data-cluster representations for all views via a mutual consensus loss function term to coordinate all views to reach clustering agreement, while structuring such representation with laplacian regularizer to encode the local manifold structure for each view.

Unlike the traditional LRR to achieve the common data similarity by all views, we propose to learn a variety of factorized low-rank data-cluster representations for different views to preserve the flexible local manifold structure while achieving the data cluster structure for each view, upon which, the consensus loss term is imposed to achieve the multi-view consensus, leading to our iterative views agreement in the next section.

\subsection{The Objective Function with structured low-rank Matrix factorized representation}
We propose the objective function with structured low-rank representation $U_i$ for each view \eg the $i^{th}$ view with factorized low-rank via Eq. \eqref{eq:reform} data-clustered representation via Eq.\eqref{eq:ui}. Then we have the following
\begin{equation}\label{eq:object}
\begin{aligned}
& \min_{U_i, E_i (i \in V)} \sum_{i \in V} ( \underbrace{\frac{1}{2}||U_i||_F^2}_\text{minimize $||Z_i||_*$ via Eq.\eqref{eq:reform}} + \underbrace{\lambda_{1} ||E_i||_1}_\text{noise and corruption robustness}\\
& + \underbrace{\lambda_{2}\textmd{Tr}(U_i^TL_iU_i)}_\text{Graph Structured Regularization} + \underbrace{\frac{\beta}{2}\sum_{j \in V, j \neq i}||U_i - U_j||^2_F)}_\text{Views-agreement}\\
&~~~\textmd{s.t.}~~~~ i = 1, \ldots, V, ~~X_i = X_iU_iU_i^T + E_i, U_i \geq 0,
\end{aligned}
\end{equation}
where
\begin{itemize}
\item $U_i \in \mathbb{R}^{n \times d}$ denotes the factorized low-rank data-cluster representation of $X_i$ for the $i^{th}$ view. $\textmd{Tr}(U_i^TL_iU_i)$ makes $U_i$ to be structured with local manifold structure for the $i^{th}$ view. $||E_i||_1$ is responsible for possible noise with $X_i$. $\lambda_{1}, \lambda_{2}, \beta$ are all trade-off parameters.

\item One reasonable assumption hold by a lot of multi-view clustering research \cite{sdm13,NIPS11,M-V-C,icml11} is that all the views should share the similar underlying clustering structure. $\sum_{i,j \in V}||U_i - U_j||_F^2$ aims to achieve the views-agreement regarding the factorized low-rank representations $U_i$ from all $|V|$ views; unlike the traditional LRR method to enforce an identical representation, we construct different $U_i$ for each view, then further minimize their divergence to generate a view-agreement.

\item $U_i \geq 0$ is a non-negative constraint, through $X_i = X_iZ_i + E_i = X_iU_iU_i^T + E_i$ for the $i^{th}$ view.
\end{itemize}
Eq.\eqref{eq:object} is non-convex, we hence alternately optimize each variable while fixing the others; that is, updating all the $U_i$ and $E_i(i \in \{1,\ldots,V\})$ in an alternative way until the convergence is reached. As solving all the $\{U_i, E_i\} (i \in V)$ pairs shares the similar optimization strategy, only the $i^{th}$ view is presented. To this end, we introduce two auxiliary variables $D_i$ and $G_i$, then solving the Eq.\eqref{eq:object} with respect to $U_i$, $E_i$, $D_i$ and $G_i$ that can be written as follows
\begin{equation}\label{eq:variation}
\begin{aligned}
& \min_{U_i, E_i, D_i, G_i} \frac{1}{2}||U_i||_F^2 + \lambda_{1}||E_i||_1 \\
& + \lambda_{2}\textmd{Tr}(U_iL_iU_i^T) + \frac{\beta}{2}\sum_{j \in V, j \neq i}||U_i - U_j||^2_F \\
& ~~\textmd{s.t.}~ X_i = D_iU_i^T + E_i, D_i = X_iU_i, G_i = U_i, G_i \geq 0,
\end{aligned}
\end{equation}
where $D_i \in \mathbb{R}^{d_i \times d}$, we will show the intuition for the auxiliary variable relationship $D_i = X_iU_i$ by introducing the augmented lagrangian function based on Eq.\eqref{eq:variation} below
\begin{equation}\label{eq:argumented}
\begin{aligned}
& \mathcal{L}(U_i, E_i, D_i, G_i, K_1^{i}, K_2^{i}, K_3^{i}) \\
& = \frac{1}{2}||U_i||_F^2 + \lambda_{1}||E_i||_1 + \lambda_{2}\textmd{Tr}(U_i^TL_iU_i)\\
& + \frac{\beta}{2}\sum_{j \in V, j \neq i}||U_i - U_j||^2_F + \langle K_1^{i}, X_i - D_iU_i^T - E_i \rangle \\
& + \langle K_2^{i}, U_i - G_i \rangle + \langle K_3^{i}, D_i - X_iU_i \rangle \\
& + \frac{\mu}{2}(||X_i - D_iU_i^T - E_i||_F^2 + ||U_i - G_i||_F^2 + ||D_i - X_iU_i||_F^2),
\end{aligned}
\end{equation}
where $K_1^{i} \in \mathbb{R}^{d_i \times n}$, $K_2^{i} \in \mathbb{R}^{n \times d}$ and $K_3^{i} \in \mathbb{R}^{d_i \times d}$ are Lagrange multipliers, $\mu > 0$ is a penalty parameter.

From Eq.\eqref{eq:argumented}, we can easily show the intuition on $D_i = X_iU_i$, that is,
\begin{itemize}
\item minimizing $||X_i - D_iU_i^T - E_i||_F^2$ w.r.t. $D_i$ is similar as dictionary learning, while pop out the $U_i^T$ as corresponding representations learning, both of them reconstruct the $X_i$ for the $i^{th}$ view. Besides the above intuition, it is quite simple to optimize only single $U_i^T$ by merging the other into $D_i$.
\end{itemize}

\section{Optimization Strategy}
We minimize Eq.\eqref{eq:argumented} by updating each variable while fixing the others.

\subsection{Solve $U_i$}
Minimizing $U_i$ is to resolve Eq.\eqref{eq:Ui}
\begin{equation}\label{eq:Ui}
\begin{aligned}
& \mathcal{L}_1 = \frac{1}{2}||U_i||_F^2 + \lambda_{2}\textmd{Tr}(U_i^TL_iU_i)\\
& + \frac{\beta}{2}\sum_{j \in V, j \neq i}||U_i - U_j||^2_F + \langle K_1^{i}, X_i - D_iU_i^T - E_i \rangle \\
& + \langle K_2^{i}, U_i - G_i \rangle + \langle K_3^{i}, D_i - X_iU_i \rangle \\
& + \frac{\mu}{2}(||X_i - D_iU_i^T - E_i||_F^2 + ||U_i - G_i||_F^2 + ||D_i - X_iU_i||_F^2),
\end{aligned}
\end{equation}

We set the derivative of Eq.\eqref{eq:Ui} w.r.t. $U_i$ to be the zero matrix, which yields the Eq.\eqref{eq:closeUi} below
\begin{equation}\label{eq:closeUi}
\begin{aligned}
& \frac{\partial \mathcal{L}_1}{\partial U_i} = U_i + 2\lambda_2L_iU_i + \beta\sum_{j \in V, j \neq i}(U_i - U_j) \\
& ~~~~~~- (K_1^i)^TD_i + K_2^i - X_i^TK_3^i \\
& ~~~~~~+ \mu U_iD_i^TD_i + \mu E_i^TD_i + \mu (U_i - G_i) \\
& ~~~~~~- \mu X_i^TX_iU_i = \mathbf{0},
\end{aligned}
\end{equation}
where $\mathbf{0} \in \mathbb{R}^{n \times d}$ shares the same size as $U_i$. Rearranging the other terms further yields the following
\begin{equation}\label{eq:closedUUi}
U_i = \underbrace{(2\lambda_2L_i + (1 + \beta(|V|-1) + \mu)I_n - \mu X_i^TX_i)^{-1}}_\text{with computational complexity $\mathcal{O}(n^3)$} S,
\end{equation}
\begin{equation}
\begin{aligned}
& S = \sum_{j \in V, j \neq i}U_j + ((K_1^i)^T - \mu U_iD_i^T - \mu E_i^T) D_i \\
& ~~~~~~+ X_i^TK_3^i + \mu X_i^TX_iU_i
\end{aligned}
\end{equation}
The bottleneck of computing Eq.\eqref{eq:closedUUi} lies in the inverse matrix computation over the matrix of the size $\mathbb{R}^{n \times n}$ causing the computational complexity $\mathcal{O}(n^3)$, which is computationally prohibitive provided that $n$ is large. Therefore, we turn to update each row of $U_i$; without loss of generality, we present the derivative with respect to $U_i(l,\cdot)$ as
\begin{equation}\label{eq:U_ik}
\begin{aligned}
& U_i(l,\cdot) + U_i(l,\cdot)\left(\sum_{k = 1}^{n} (2\lambda_2L_i(k,l) - \mu (X_i^TX_i)(k,l))\right) \\
& + (K_1^i)^T(l,\cdot)D_i + \mu U_i(l,\cdot)D_i^TD_i + K_2^i(l,\cdot) - X_i^T(l,\cdot)K_3^i  \\
& +\beta\sum_{j \in V, j \neq i}(U_i(l,\cdot) - U_j(l,\cdot)) \\
& + \mu\left(U_i(l,\cdot) + E_i^T(l,\cdot)D_i - G_i(l,\cdot)\right) = \mathbf{0},
\end{aligned}
\end{equation}
where $\mathbf{0} \in \mathbb{R}^{d}$ denotes the vector of the size $d$ with all entries to be 0, $U_i(l,\cdot) \in \mathbb{R}^{d}$ represents the $l^{th}$ row of $U_i$; we rearrange the terms to yield the following
\begin{equation}\label{eq:uil}\small
\begin{aligned}
& U_i(l,\cdot) \\
& = \left(T_i^l  + \beta\underbrace{\sum_{j \neq i, j \in V}U_j(l,\cdot)}_\text{Influences from other views} \right) \\
&\underbrace{\left((1 + \mu + \sum_{k = 1}^{n}(2\lambda_2L_i(k,l) - \mu (X_i^TX_i)(k,l)))I_d + D_i^TD_i\right)^{-1}}_\text{with computational complexity $\mathcal{O}(d^3)$}
\end{aligned}
\end{equation}
where
\begin{equation}
\begin{aligned}
& T_i^l = X_i^T(l,\cdot)K_3^i + \mu \left(G_i(l,\cdot) - E_i^T(l,\cdot)D_i\right)\\ \nonumber
& ~~~- K_2^i(l,\cdot) - (K_1^i)^T(l,\cdot)D_i \nonumber
\end{aligned}
\end{equation}

$I_d \in \mathbb{R}^{d \times d}$ is the identity matrix.
\subsubsection{\textbf{Complexity discussion for the row updating strategy for $U_i$}}
Unlike the closed form regarding $U_i$, it is apparent that the major computational complexity lies in the inverse matrix computation over the size of $\mathbb{R}^{d \times d}$, which leads to $\mathcal{O}(d^3)$ according to Eq.\eqref{eq:uil}, which is much smaller than $\mathcal{O}(n^3)$. Besides, as $d$ is set as the cluster number across all views; moreover, aforementioned, it should be less than the inherent rank of $X_i$, and hence a small value. Upon the above facts, it is tremendously efficient via $\mathcal{O}(d^3)$ to sequentially update each row of $U_i$.
\subsubsection{\textbf{Intuitions for views agreement}}
The iterative views clustering agreement can be immediately captured via the terms underlined in Eq.\eqref{eq:uil}. Specifically, during each iteration, the $U_i(l,\cdot)$ is updated via the influence from others view, while served as the constraint to generate $U_j(l,\cdot) (j \neq i)$, the divergence among all $U_i(l,\cdot)$ is decreased gradually towards an agreement for all views, such process repeats until the convergence is reached.

Unlike the existing LRR method by directly imposing the common representation, our iterative multi-view agreement can better preserve the flexible manifold structure for each view meanwhile achieve the multi-view agreement, which will be critical to final multi-view spectral clustering. \\
\textbf{Remark.} After the whole $U_i$ is updated for the $i^{th}$ view, we simply perform a K-means clustering over it to assign each data object to one cluster exclusively. Then normalized each column of $U_i$ to form an orthonormal matrix.

\subsection{Solve $D_i$}
The optimization process regarding $D_i$ is equivalent to the following
\begin{equation}\label{eq:dii}
\begin{aligned}
& \min_{D_i}<K_i^i, X_i - D_iU_i^T - E_i> + <K_3^i, D_i - X_iU_i> \\
& + \frac{\mu}{2}\left(||X_i - D_iU_i - E_i||_F^2 + ||D_i - X_iU_i||_F^2\right)
\end{aligned}
\end{equation}
We get the derivative with respect to $D_i$, then it yields the following closed form updating rule
\begin{equation}\label{eq:diupdate}
\begin{aligned}
& D_i = \\
& \left(K_1^iU_i - K_3^i + \mu(2X_i - E_i)U_i\right)\frac{\left(I_d + U_i^TU_i\right)^{-1}}{\mu},
\end{aligned}
\end{equation}
where the major computational complexity lies in the inverse computation over matrix $(I_d + U_i^TU_i) \in \mathbb{R}^{d \times d}$, resulting into $\mathcal{O}(d^3)$, as aforementioned, that is the same as updating each row of $U_i$, and hence quite efficient.

\subsection{Solve $E_i$}
it is equivalent to solving the following:
\begin{equation}\label{eq:Ei}
\min_{E_i} \lambda_1 ||E_i||_1 + \frac{\mu}{2}||E_i - (X_i - D_iU_i^T + \frac{1}{\mu}K_1^{i})||_F^2,
\end{equation}
where the following closed form solution can be yielded for $E_i$ according to \cite{CaiSIAMJ08}
\begin{equation}\label{eq:EiS}
E_i = S_{\frac{\lambda_1}{\mu}}(X_i - D_iU_i^T + \frac{1}{\mu}K_1^{i})
\end{equation}

\subsection{Solve $G_i$}
It is equivalent for the following:
\begin{equation}\label{eq:Gi}
<K_2^i,U_i - G_i> + \frac{\mu}{2}||G_i - U_i||_F^2
\end{equation}
where the following closed form solution of $G_i$ can be derived as
\begin{equation}\label{eq:GiS}
G_i = U_i + \frac{K_2^i}{\mu}
\end{equation}
\begin{algorithm}[hbt]
\KwIn{$X_i (i = 1,\ldots, V), d, \lambda_1, \lambda_2, \beta$ }
\KwOut{$U_i, D_i, E_i, G_i (i \in V)$ }
\textbf{Initialize}: $U_i[0], L_i (i = 1, \dots,V)$ computation, set all entries of $K_1^{i}[0],G_i[0], K_2^{i}[0]$ to be 0, initialize $E_i[0]$ with sparse noise as 20\% entries corrupted with uniformly distributed noise over [-5,5], $\mu[0]=10^{-3}$, $\epsilon_1=10^{-3}$, $\epsilon_2=10^{-1}$\\
$k = 0$\\
\For {$i \in V$}{
\textbf{Solve $U_i$}:\\
Sequentially update each row of $U_i$ according to Eq.\eqref{eq:uil}.\\
Orthonormalized each column of $U_i$.\\
\textbf{Update $E_i$}:\\
$E_i[k + 1] = S_{\frac{\lambda_1}{\mu[k]}}(X_i - D_iU_i^T[k] + \frac{1}{\mu[k]}K_1^{i}[k])$\\
\textbf{Update $G_i$}:\\
$G_i[k + 1] = U_i[k] + \frac{K_2^i[k]}{\mu[k]}$\\
\textbf{Update $K_1^{i}$, $K_2^{i}$, $K_3^{i}$ and $\mu$}:\\
$K_1^{i}[k + 1] = K_1^{i}[k] + \mu(X_i - D_iU_i^T[k] - E_i[k])$\\
$K_2^{i}[k + 1] = K_2^{i}[k] + \mu(U_i[k] - G_i[k])$\\
$K_3^{i}[k + 1] = K_3^{i}[k] + \mu(D_i[k] - X_iU_i[k])$\\
\textbf{Update $\mu$} according to \cite{LinNIPS2011}\\
\textbf{whether converged}\\
\If {$||X_i - D_iU_i^T[k + 1] - E_i[k+1]|| /||X_i|| < \epsilon_1$ and  \\
\textmd{max}$\{\xi ||U_i[k + 1] - U_i[k]||, \mu[k]||G_i[k + 1] - G_i[k]||, \mu[k]||E_i[k+1] - E_i[k]||\} < \epsilon_2$}
{Remove the $i^{th}$ view from the view set as $V = V - i$\\
$U_i[N] = U_i[k + 1]$, s.t. $N$ is any positive integer.\\
}
\Else{$k = k + 1$}
}
\textbf{Return} $U_i[k + 1]$, $D_i[k + 1]$, $E_i[k + 1]$, $G_i[k + 1]$ ($i=1,\ldots,V$)
\caption{\small Alternating optimization strategy for Eq.\eqref{eq:object}.}\label{alg:algorithm1}\small
\end{algorithm}
\subsection{Updating $K_1^{i}$, $K_2^{i}$, $K_3^{i}$ and $\mu$}
We update Lagrange multipliers $K_1^{i}$, $K_2^{i}$ and $K_3^{i}$ via

\begin{equation}\label{eq:K1s}
K_1^{i} = K_1^{i} + \mu(X_i - D_iU_i - E_i)
\end{equation}
\begin{equation}\label{eq:K2s}
K_2^{i} = K_2^{i} + \mu(U_i - G_i)
\end{equation}
\begin{equation}\label{eq:K3s}
K_3^{i} = K_3^{i} + \mu(D_i - X_iU_i)
\end{equation}

Following \cite{ijcai16}, $\mu$ is tuned using the adaptive updating strategy \cite{LinNIPS2011} to yield a faster convergence. The optimization strategy alternatively updates each variable while fixing others until the convergence, which is summarized by Algorithm ~\ref{alg:algorithm1}.


\subsection{Notes regarding Algorithm ~\ref{alg:algorithm1}}
It is worthwhile to highlight some critical notes regarding the Algorithm ~\ref{alg:algorithm1} below

\begin{itemize}

\item We initialize the $U_i[0] \in \mathbb{R}^{n \times d}$ for all views, such that each entry of $U_i[0]$ represents similarity between each data object and one of the $d$ anchors (cluster representatives), which can be seen as the centers from the clusters generated from the k-means or spectral clustering.

\item For our initialization, we adopt the spectral clustering outcome with the clustering number to be $d$, where the similarity matrix is calculated via the original $X_i$ feature representation within each view, then the $U_i[0](i,j)$ entry \ie the similarity between the $i^{th}$ data object and the $j^{th}$ anchor is yielded via Eq.\eqref{eq:gaussian}. The laplacian matrix $L_i (i = 1,\ldots,V)$ are computed once offline also within the original $X_i$ feature representation.

\item More importantly, we set the identical value of $d$(the cluster number) to the column size of $U_i[0](i = 1,\cdot\cdot,V) \in \mathbb{R}^{n \times d}$ for all the views. We remark that the above initial setting for $U_i[0]$ with the same $d$ is reasonable, as stated before all the views should share the similar underlying data clustering structure. This fact also implies that the initialized $U_i[0]$ is reasonably not divergent a lot among all views.

\end{itemize}

\subsection{Convergence discussion}
Often, the above alternating minimization strategy can be seen as the coordinate descent method. According to \cite{nonlinearconv}, the sequences $(U_i,D_i,E_i,G_i)$ above will eventually converge to a stationary point. However, we are not sure whether the converged stationary point is a global optimum, as it is not jointly convex to all the variables above.

\subsection{Clustering}
Following \cite{ijcai16}, once the converged $U_i(i = 1,\ldots,V)$ are ready, all column vectors of $U_i(i = 1,\ldots,V)$ while set small entries under given threshold $\tau$ to be 0. Afterwards, the similarity matrix
for the $i^{th}$ view between the $j^{th}$ and $k^{th}$ data objects as
\begin{equation}\label{eq:all-view-simi}
W_i (j,k) =(U_iU_i^T)(j,k)
\end{equation}
Following \cite{ijcai16}, The final data similarity matrix can be defined as
\begin{equation}\label{eq:simi-all}
W = \frac{\sum_{i}^{V}W_i}{|V|}
\end{equation}
The clustering is carried out against $W$ via Eq.\eqref{eq:simi-all} to yield final outcome of $d$ data groups.

\section{Experiments}\label{sec:exp}
We adopt the data sets mentioned in \cite{ijcai16} below:

\begin{itemize}
\item \underline{UCI handwritten Digit set}\footnote{http://archive.ics.uci.edu/ml/datasets/Multiple+Features}: consists of features for hand-written digits (0-9), with 6 features and contains 2000 samples with 200 in each category. Analogous to \cite{LinNIPS2011,ijcai16}, we choose two views as 76 Fourier coefficients  (FC) of the character shapes and the 216 profile correlations.

\item \underline{Animal with Attribute} (AwA)\footnote{http://attributes.kyb.tuebingen.mpg.de}: consists of 50 kinds of animals described by 6 features (views): Color histogram ( CQ, 2688-dim),  local self-similarity (LSS, 2000-dim),  pyramid HOG (PHOG, 252-dim), SIFT (2000-dim), Color SIFT (RGSIFT, 2000-dim), and SURF (2000-dim). Following \cite{ijcai16}, 80 images for each category and get 4000 images in total.

\item \underline{NUS-WIDE-Object (NUS)} \cite{NUS-Wide}: 30000 images from 31 categories. 5 views are adopted using 5 features as provided by the website \footnote{lms.comp.nus.edu.sg/research/NUS-WIDE.html}: 65-dimensional color histogram (CH), 226-dimensional color moments (CM), 145-dimensional color correlation (CORR), 74-dimensional edge estimation (EDH), and 129-dimensional wavelet texture (WT).

\item \underline{PASCAL VOC 2012}\footnote{http://host.robots.ox.ac.uk/pascal/VOC/voc2012/}: we select  20 categories with 11530 images, two views are constructed with Color features (1500-dim) and HOG features (250 dim). Among them, 5600 images are selected by removing the images with multiple categories.
\end{itemize}

We summarize the above throughout Table \ref{table:dataset}.
\begin{table}[t]
\tiny
\caption{Data sets.}
\begin{center}
\begin{tabular}{ccccc}
\hline
Features & UCI digits & AwA & NUS & VOC\\
\hline
1 & FC (76) & CQ (2688) & CH(65) & Color (1500)\\
2 & PC (216) & LSS (2000) & CM(226) & HOG (250)\\
3 & - & PHOG (252)& CORR(145) & -\\
4 & - &SIFT(2000) & EDH(74) & - \\
5 & -& RGSIFT(2000) & WT(129) & - \\
6 & - & SURF(2000) & - & - \\
\hline
\# of data & 2000 & 4000 & 26315 & 5600\\
\# of classes & 10 & 50 & 31 & 20\\
\hline
\end{tabular}
\label{table:dataset}
\end{center}
\end{table}

\subsection{Baselines}
The following state-of-the-art baselines used in \cite{ijcai16} are compared:

\begin{itemize}
\item \textbf{MFMSC}: concatenating multi-features to be the multi-view representation for similarity matrix, the spectral clustering is then conducted \cite{cvpr10}.

\item Multi-feature representation similarity aggregation for spectral clustering (\textbf{MAASC}) \cite{CVPR12}.

\item Canonical Correlation Analysis (CCA) model (\textbf{CCAMSC}) \cite{ICML09}: Projecting multi-view data into a common subspace, then perform spectral clustering.

\item Co-regularized multi-view spectral clustering (\textbf{CoMVSC}) \cite{NIPS11}: It regularizes the eigenvectors of view-dependent graph laplacians and achieve consensus clusters across views.

\item \textbf{Co-training} \cite{icml11}: Alternately modify one view's Laplacian eigenspace  by learning from the other views 's eigenspace, the spectral clustering is then conducted.

\item Robust Low-Rank Representation method (\textbf{RLRR}) \cite{RMVSC},  after obtaining the data similarity matrix, upon which, the spectral clustering is performed to be the final multi-view spectral clustering result.

\item Low-rank Representation with Graph laplacian (\textbf{LRRGL}) \cite{ijcai16} regularizer over the non-factorized low-rank representations, with each of which corresponds to one view to preserve the individual manifold structure, while iteratively boost all these low-rank representations to reach agreement. The final multi-view spectral clustering is performed upon the similarity representations
\end{itemize}

\subsection{Experimental Settings and Parameters Study}

 We implement these competitors under the experimental  setting as mentioned in \cite{ijcai16}. Following \cite{ijcai16},  $\sigma$  in Eq.\eqref{eq:gaussian} is learned via \cite{nips04}, and
$s=20$ to construct $s$-nearest neighbors for Eq.\eqref{eq:gaussian}.
We adopt two standard metrics: clustering accuracy (\textsf{ACC}) and normalized mutual information (\textsf{NMI}) as the metric defined as Eq.\eqref{eq:acc}
\begin{equation}\label{eq:acc}
\textsf{ACC} = \frac{\sum_{i=1}^n \delta(\textmd{map}(r_i),l_i)}{n},
\end{equation}
where $r_i$ denotes the cluster label of $x_i$, and $l_i$ denotes the true class label, $n$ is the total number of images, $\delta(x,y)$ is the function that equals one if $x=y$ and equals zero otherwise, and $\textmd{map}(r_i)$ is the permutation mapping function that maps each cluster label $r_i$ to the equivalent label from the database. Meanwhile the \textsf{NMI} is formulated below
\begin{equation}\label{eq:NMI}
\textsf{NMI} = \frac{\sum_{i=1}^c \sum_{j=1}^c n_{i,j} \log \frac{n_{i,j}}{n_i \hat{n}_j}}{\sqrt{(\sum_{i=1}^c n_i \log \frac{n_i}{n}) (\sum_{j=1}^c \hat{n}_j \log \frac{\hat{n}_j}{n})}},
\end{equation}
where $n_i$ is the sample number in cluster $C_i$ ($1\leqslant i \leqslant c$), $\hat{n}_j$ is the sample number from class $L_j$ ($1 \leqslant j \leqslant c$), and $n_{i,j}$ denotes the sample number in the intersection between $C_i$ and $L_j$.\\
\textbf{Remark.} Following \cite{ijcai16}, \emph{we repeated the running 10 times, and their averaged mean value for multi-view spectral clustering for all methods is reported. For each method including ours, we input the clustering number as the number of ground-truth classes from all data sets.}\\
\textbf{\underline{Feature noise modeling for robustness}}: Following \cite{CV155,ijcai16}, 20\% feature elements are corrupted with uniform distribution over the range [5,-5], which is consistent to the practical setting while matching with \textbf{LRRGL},\textbf{RLRR} and our method.

Following \cite{ijcai16}, We set $\lambda_1=2$ in Eq.\eqref{eq:object} for sparse noise term.  We test \textsf{ACC} and \textsf{NMI} over different value of $\lambda_2$ and $\beta$ in Eq.\eqref{eq:object} in the next subsection.

\subsection{Validation over factorized low-rank latent data-cluster representation}
First, we will would like to validate our method regarding the multi-graph regularization and iterative views agreement over factorized latent data-cluster representation.

\begin{figure*}[t]
\begin{center}
\begin{tabular}{ccc}
\includegraphics[width=6cm]{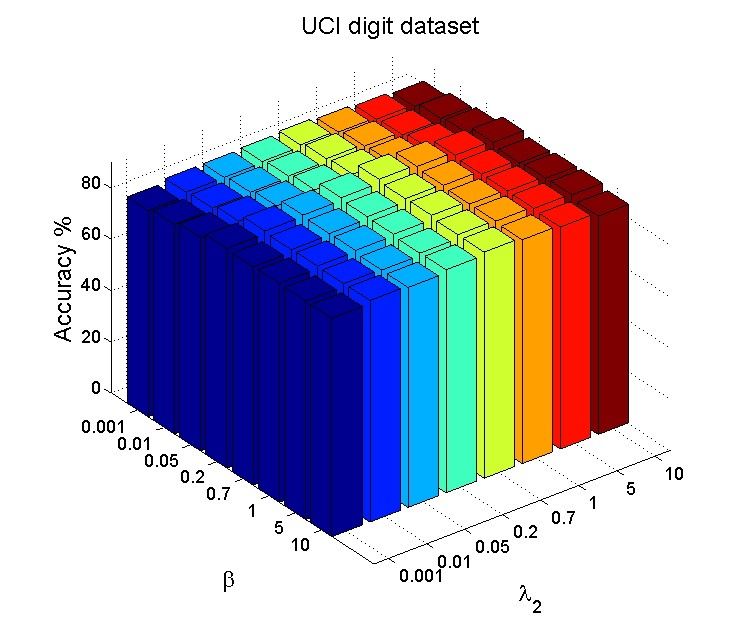}&
\includegraphics[width=6cm]{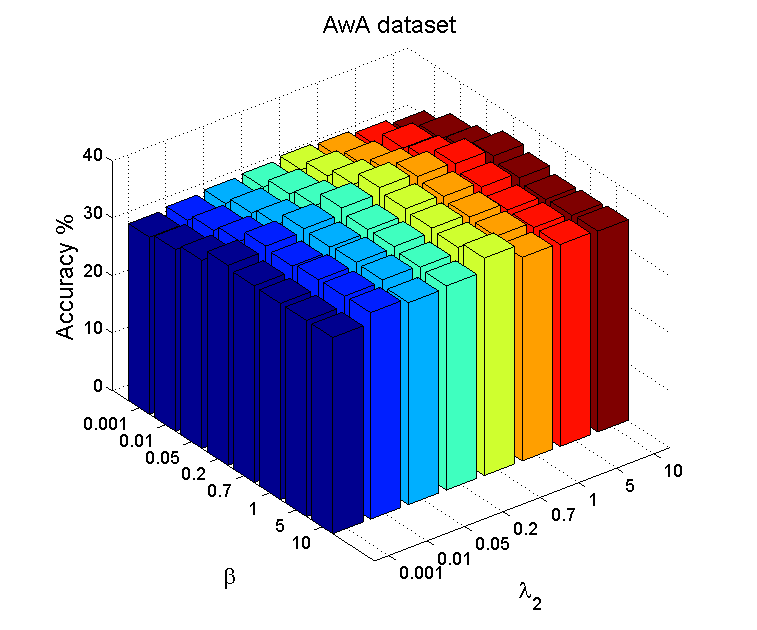}&
\includegraphics[width=6cm]{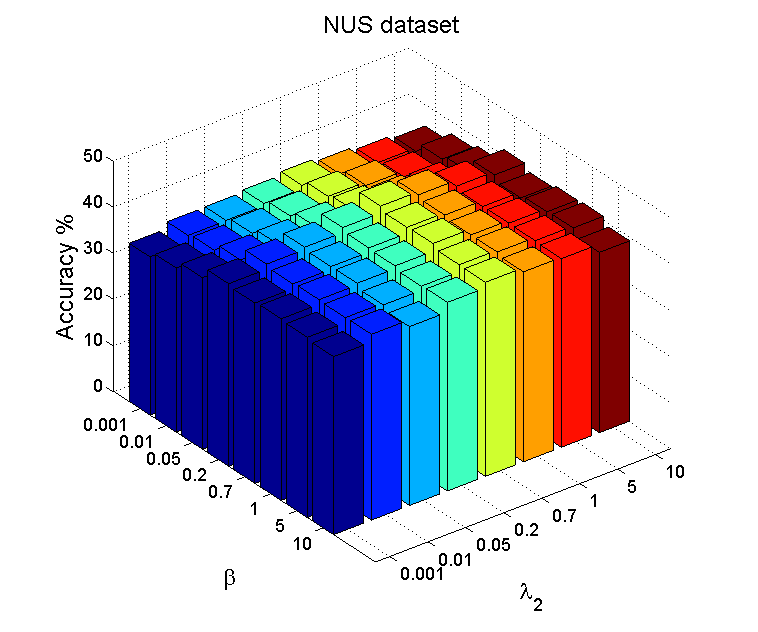}\\
(a)&(b)&(c)
\end{tabular}
\caption{Study over $\lambda_2$ and $\beta$ over latent factorized data-cluster representation on three datasets.}\label{fig:parameter}
\end{center}
\end{figure*}

Following \cite{ijcai16}, we test $\lambda_2$ and $\beta$ within the interval [0.001,10], with one parameter while fixing the value of the other parameter, the \textsf{ACC} results are shown in Fig. \ref{fig:parameter}, where we have
\begin{itemize}
\item Increasing $\beta$ will improve the performance, and vice versa; that is, increasing $\lambda_2$ will improve the performance.
\item The clustering metric \textsf{ACC} increases when both $\lambda_2$ and $\beta$ increase.
\end{itemize}
Based on the above,  we choose a balance pair values: $\lambda_2=0.7$ and $\beta=0.2$ for our method.

\subsection{Results}
\begin{table}[t]
\caption{\textsf{ACC} results.}
\begin{center}
\tiny
\begin{tabular}{ccccc}
\hline\hline
\textsf{ACC} (\%) & UCI digits & AwA & NUS & VOC\\
\hline
\textbf{MFMSC} \cite{cvpr10} & 43.81 & 17.13& 22.81 & 12.98\\
\textbf{MAASC} \cite{CVPR12} & 51.74 & 19.44& 25.13 & 13.64\\
\textbf{CCAMSC}\cite{ICML09} & 73.24 & 24.04& 27.56 & 12.05\\
\textbf{CoMVSC}\cite{NIPS11} & 80.27 & 29.93 & 33.63 & 14.03\\
\textbf{Co-training}\cite{icml11} & 79.22 & 29.06 & 34.25 & 14.92\\
\textbf{RLRR}\cite{RMVSC} & 83.67 & 31.49 & 35.27 & 17.13\\
\textbf{LRRGL}\cite{ijcai16} & 86.39 & 37.22 & 41.02 & 18.07\\
\hline
\textbf{Ours} & \textbf{89.64} & \textbf{41.76} & \textbf{43.14} & \textbf{18.85}\\
\hline
\end{tabular}
\end{center}
\label{table:acc}
\end{table}

\begin{table}[t]
\caption{\textsf{NMI} results.}
\begin{center}
\tiny
\begin{tabular}{ccccc}
\hline\hline
\textsf{NMI} (\%) & UCI digits & AwA & NUS & VOC\\
\hline
\textbf{MFMSC} \cite{cvpr10} & 41.57 & 11.48 & 12.21 & 9.16\\
\textbf{MAASC} \cite{CVPR12} & 47.85 & 12.93& 11.86 & 9.84\\
\textbf{CCAMSC}\cite{ICML09} & 56.51 & 15.62& 14.56 & 8.42\\
\textbf{CoMVSC}\cite{NIPS11} & 63.82 & 17.30 & 7.07 & 9.97\\
\textbf{Co-training}\cite{icml11} & 62.07 & 18.05 & 8.10 & 10.88\\
\textbf{RLRR}\cite{RMVSC} & 81.20 & 25.57 & 18.29 & 11.65\\
\textbf{LRRGL}\cite{ijcai16} & 85.45 & 31.74 & 20.61 & 12.03\\
\hline
\textbf{Ours} & \textbf{87.81} & \textbf{34.03} & \textbf{23.43} & \textbf{12.97}\\
\hline
\end{tabular}
\end{center}
\label{table:nmi}
\end{table}

According to Table \ref{table:acc} and Table \ref{table:nmi},  the following identification can be drawn, note that we mainly deliver the analysis between our method and \textbf{LRRGL}, as the analysis over other competitors have been detailed in \cite{ijcai16}.

\begin{figure*}[t]
\centering
\begin{tabular}{ccc}
\includegraphics[width=5cm,height=5cm]{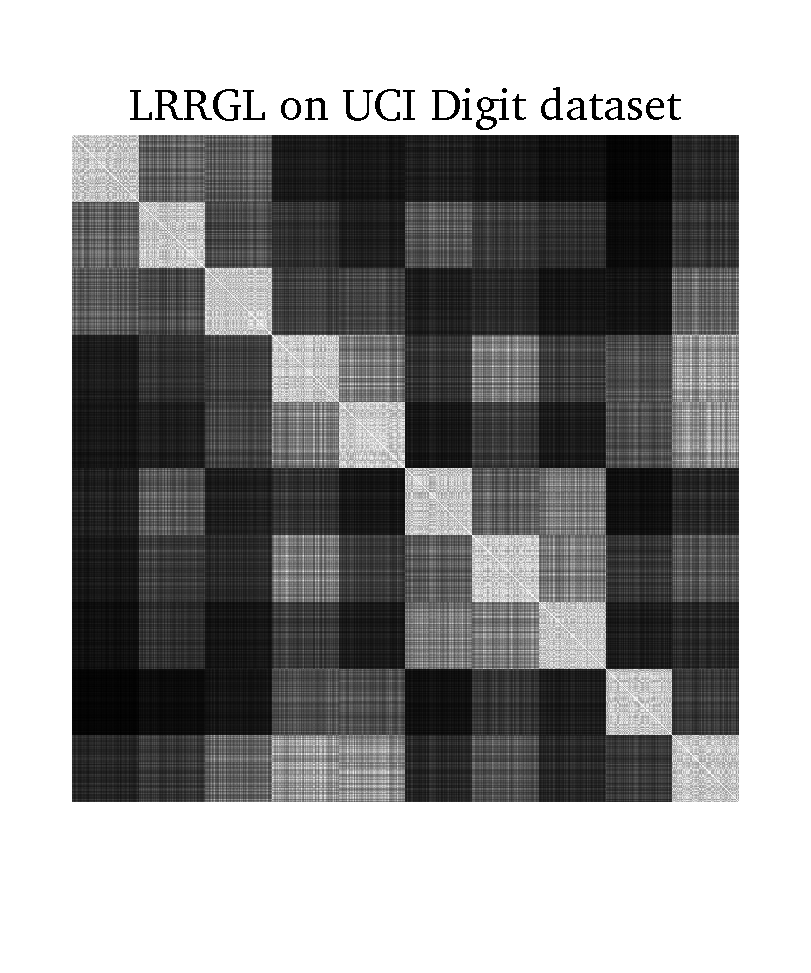}&
\includegraphics[width=5cm,height=5cm]{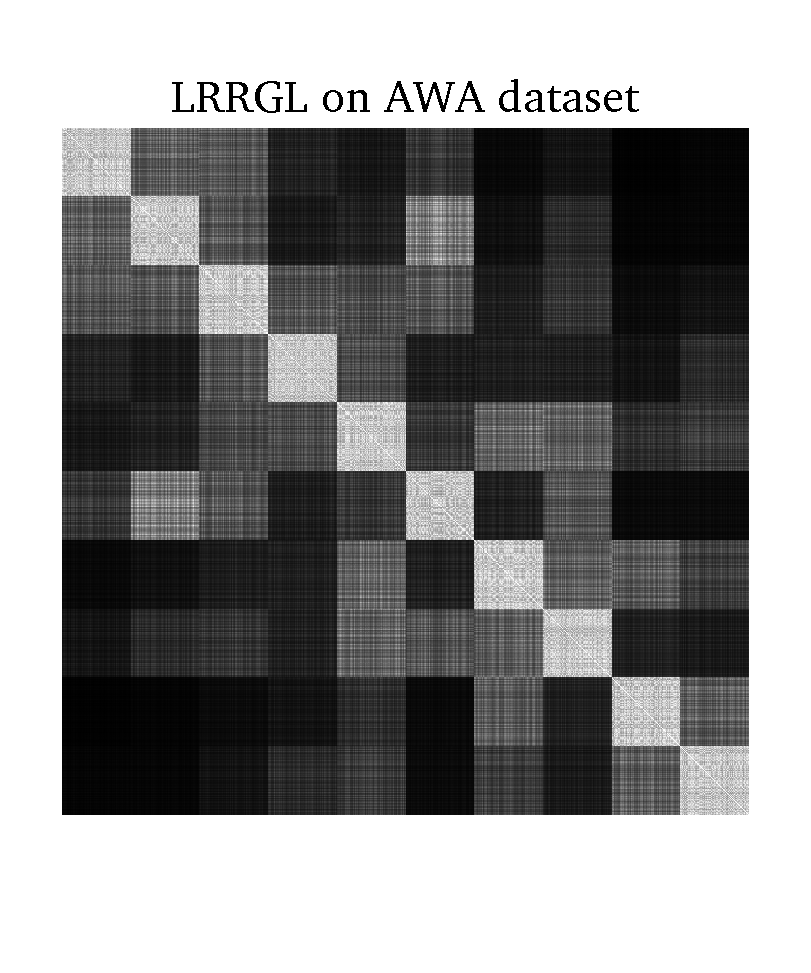}&
\includegraphics[width=5cm,height=5cm]{ijcai_nus}\\
(a)&(b)&(c)\\
\includegraphics[width=5cm,height=5cm]{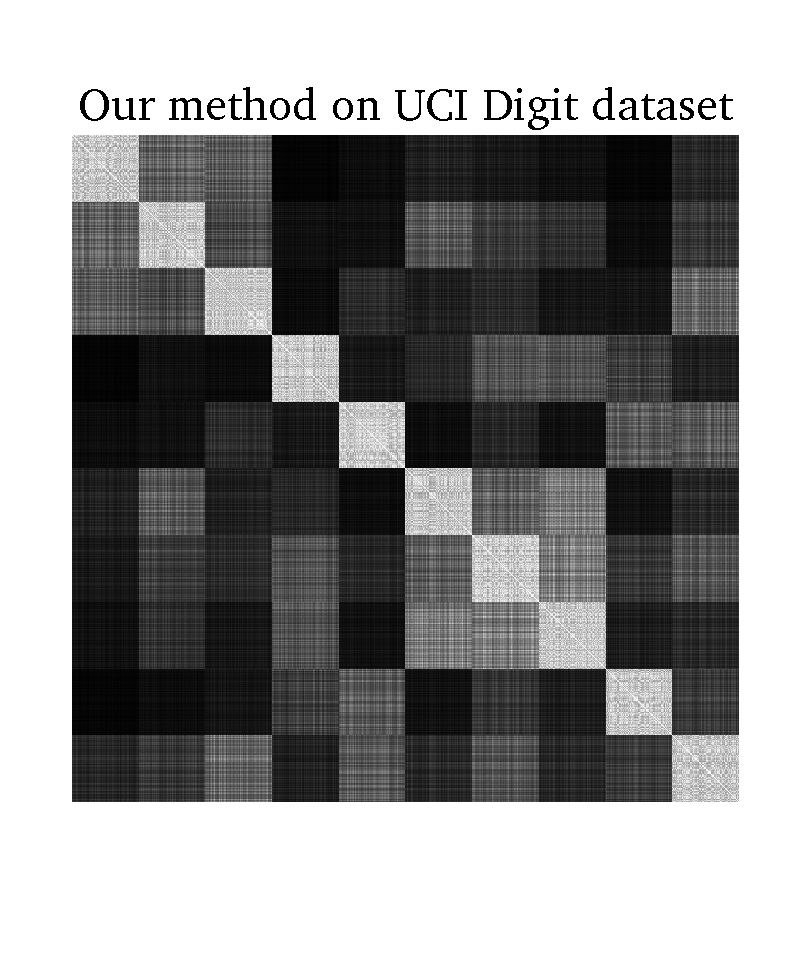}&
\includegraphics[width=5cm,height=5cm]{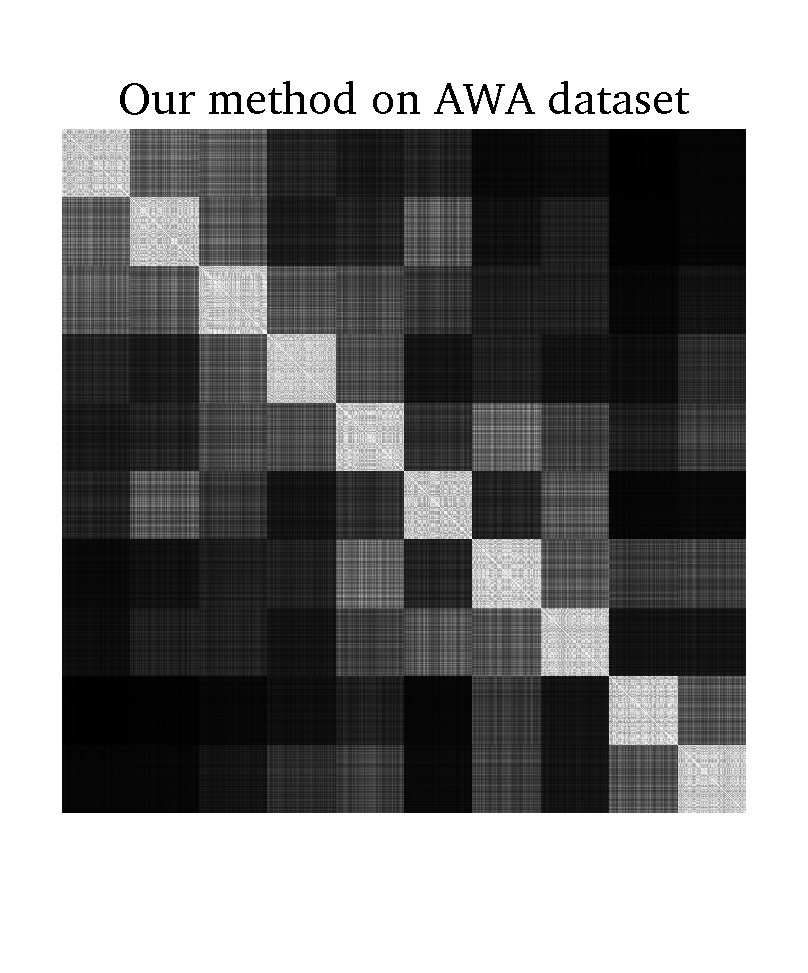}&
\includegraphics[width=5cm,height=5cm]{structured_cluster_nus}\\
(d)&(e)&(f)
\end{tabular}
\caption{Recovered multi-view based consensus affinity matrix over both our proposed method and \textbf{LRRGL} on three multi-view data sets with noise corruption. For UCI digit dataset, we plot the affinity matrix over all data sample. For AWA and NUS datasets, we randomly select 10 classes, where 80 samples are randomly selected for each of them. The 10 diagonal block represents the data samples within the 10 clusters w.r.t ground truth classes, where more white the color is, the ideally larger affinity value will be  to better reveal the data samples clusters within the same classes. Meanwhile, for non-diagonal blocks, the more black the color is, the ideally smaller affinity will be to reveal the data samples within different clusters w.r.t ground truth classes. }\label{fig:affinity}
\end{figure*}

\begin{itemize}
\item First, our method outperforms \textbf{LRRGL}, implying the effectiveness of the factorized latent data-cluster representation, as it can better encode the data-cluster representation for each view as well as all views. We provide more insights about that in Fig.~\ref{fig:affinity}.

\item Second, both our method and \textbf{LRRGL} outperforms the model of learning a common low-dimensional subspace among multi-view data, as indicated by \cite{ijcai16} it is incapable of encoding local graph structures within a single subspace.

\item Our method and \textbf{LRRGL} are more effective under noise corruptions than other methods.  More analysis can be referred to our conference version \cite{ijcai16}.

\item Our method achieves the best performance over PASCAL VOC 2012 under the selected two views via the tuned the parameters.
\end{itemize}

We present Fig. ~\ref{fig:affinity} to show more intuitions on why our method with the multi-view affinity matrix yielded from factorized data-cluster representation outperforms the primal similarity matrix for \textbf{LRRGL}.  For example,
\begin{itemize}
\item For UCI dataset, \ie the multi-view affinity matrix illustrated in Fig.~\ref{fig:affinity}(a) and ~\ref{fig:affinity}(d), we can see the both $4^{th}$ and $5^{th}$ diagonal blocks of our method in Fig.~\ref{fig:affinity}(d) are more whiter than those of \textbf{LRRGL} illustrated in Fig. ~\ref{fig:affinity}(a); meanwhile the surrounding non-diagonal black blocks \eg $(4,5)^{th}$ and $(5,4)^{th}$ are more black than those of textbf{LRRGL}.
\item For AwA dataset, the diagonal blocks of our method from the $2^{nd}$ to the $6^{th}$ are whiter than those of \textbf{LRRGL}, leading to a slight deeper black color over the surrounding non-diagonal blocks than \textbf{LRRGL}.
\item The similar conclusions also hold for NUS dataset, we can see the diagonal blocks from $3^{rd}$ to the $8^{th}$ of our method are more whiter than those of \textbf{LRRGL}, leading to the result that the surrounding black non-diagonal blocks of our method are more salient than those of \textbf{LRRGL}.
\end{itemize}
From the above observations, we can safely infer the advantages of the affinity matrix representation yielded by our factorized latent data-cluster representation over the primal affinity matrix of \textbf{LRRGL} for Multi-view spectral clustering.

\section{Conclusion}
In this paper, we propose to learn a clustered low-rank representation via structured matrix factorization for multi-view spectral clustering. Unlike the existing methods, we propose an iterative strategy of intuitively achieving the multi-view spectral clustering agreement by minimizing the between-view divergences in terms of the factorized latent data-clustered representation for each view. Upon that, we impose the graph Laplacian regularizer over such low-dimensional data-cluster representation, so as to adapt to the multi-view spectral clustering, as demonstrated by the extensive experiments.

The future work includes the following directions: The graph regularized low-rank embedding out-of-sample case has been researched \cite{Fei11TNN}, and will be applied for multi-view out-of-sample scenario. Unlike the pre-defined graph similarity value, inspired by \cite{FeiAAAI17}, we will simultaneously learn and achieve the consensus graph clustering result and graph structure \ie graph similarity. Besides, the latest non-parametric graph construction model \cite{FeiAAAI16} will also be incorporated for multi-view spectral clustering. The practice of our method can be improved by reducing the tuned parameters further. Upon that, we will also investigate the problem of learning the weight \cite{FeiAAAI17,IJCAI16Nie,AAAI13Cai} for each view.

{
\bibliographystyle{named}
\bibliography{ijcai17}
}
\end{document}